# Assessing the Feasibility of Early Cancer Detection Using Routine Laboratory Data: An Evaluation of Machine Learning Approaches on an Imbalanced Dataset

**Author**: Shumin Li[1]

**Affiliation:**

[1] College of Veterinary Medicine, Jilin University, Jilin, China 130062

**Corresponding author** :

Shumin Li, DVM, MSc

College of Veterinary Medicine, Jilin University

5333 Xi'an Road, Lvyuan District, Changchun, Jilin, China 130062

shuminli@jlu.edu.cn or aishanglishumin@gmail.com (preferred)

+8618100481863

**Abstract:** The development of accessible screening tools for early cancer detection in dogs represents a significant challenge in veterinary medicine. Routine laboratory data offer a promising, low-cost source for such tools, but their utility is hampered by the non-specificity of individual biomarkers and the severe class imbalance inherent in screening populations. This study assesses the feasibility of cancer risk classification using the Golden Retriever Lifetime Study (GRLS) cohort under real-world constraints, including the grouping of diverse cancer types and the inclusion of post-diagnosis samples. A comprehensive benchmark evaluation was conducted, systematically comparing 126 analytical pipelines that comprised various machine learning models, feature selection methods, and data balancing techniques. Data were partitioned at the patient level to prevent leakage. The optimal model, a Logistic Regression classifier with class weighting and recursive feature elimination, demonstrated moderate ranking ability (AUROC = 0.815; 95% CI: 0.793-0.836) but poor clinical classification performance (F1-score = 0.25, Positive Predictive Value = 0.15). While a high Negative Predictive Value (0.98) was achieved, insufficient recall (0.79) precludes its use as a reliable rule-out test. Interpretability analysis with SHapley Additive exPlanations (SHAP) revealed that predictions were driven by non-specific features like age and markers of inflammation and anemia. It is concluded that while a statistically detectable cancer signal exists in routine lab data, it is too weak and confounded for clinically reliable discrimination from normal aging or other inflammatory conditions. This work establishes a critical performance ceiling for this data modality in isolation and underscores that meaningful progress in computational veterinary oncology will require integration of multi-modal data sources.

Keywords: Veterinary Oncology, Machine Learning, Predictive Modeling, Imbalanced Data, Routine Laboratory Data, Bloodwork, Early Detection, Golden Retriever Lifetime Study (GRLS)

# 1. Background

Cancer is a leading cause of mortality in companion dogs, with an incidence that increases with age, presenting great emotional and clinical challenges [1, 2, 3]. This challenge is compounded by significant diagnostic gaps. For instance, a 2025 survey by veterinary imaging company HT Vista found that 62% of masses on dogs seen in US veterinary clinics go undiagnosed [4]. This high rate of undiagnosed cases highlights the need for scalable, cost-effective screening tools. Given that dogs also serve as valuable spontaneous models for human oncology [5], the development of such tools is a key objective that promises not only to improve therapeutic outcomes in dogs but also to yield insights that inform human medicine. In this pursuit, routine laboratory diagnostics, such as the Complete Blood Count (CBC) and serum biochemistry panels (Chem), are uniquely positioned as a potential data source. These tests are among the most frequently performed procedures in veterinary medicine, generating a massive volume of structured, longitudinal data ripe for computational analysis. The central hypothesis is that while individual parameters may be uninformative, subtle, multivariate patterns within this data may harbor a presymptomatic signature of malignancy.

The application of machine learning (ML) to such routine diagnostics has shown potential. Previous veterinary research demonstrated success in using supervised algorithms to distinguish between inflammatory bowel disease and alimentary lymphoma in cats [6]. This approach is an even more active area of research in human oncology, where models have been developed to predict lung cancer months before clinical diagnosis and to identify patients at high risk for a variety of cancers [7-9]. These successes highlight the theoretical power of using ML to detect faint biological signals in high-dimensional data.

However, developing a clinically reliable diagnostic tool faces hurdles, primarily stemming from the biological non-specificity of hematological markers. Alterations such as anemia are not pathognomonic for cancer but often reflect systemic inflammation, which is common in geriatric patients with non-neoplastic diseases. This limitation is exemplified in a study [10] on feline intestinal disease, where incorporating complete blood count and serum biochemistry data with ultrasound radiomics failed to significantly improve ML model accuracy, despite evidence from numerous other studies indicating that such data integration typically enhances performance [11,

12]. This finding aligns with a more recent large-scale human study, which concluded that while adding blood tests improved sensitivity for cancer prediction, it did not enhance the model's overall discriminatory power [13]. Compounding these biological challenges is the statistical problem of low disease prevalence. In a typical screening cohort, the vast majority of individuals are cancer-free, creating a severely imbalanced dataset. This imbalance is notorious for causing ML algorithms to develop a bias toward the majority class, further hindering accurate cancer detection.

This study utilizes data from the Morris Animal Foundation (MAF) Golden Retriever Lifetime Study (GRLS), a large prospective observational study of 3,044 purebred Golden Retrievers [14, 15]. The GRLS dataset has been leveraged in numerous canine cancer studies [16-20]. A few studies shared our interest in hematological data, and investigated them as mortality predictors [21] and their relationship with endoparasitism [22]. This rich history of use, combined with the cohort's established translational relevance to human health [14, 23], makes the GRLS an ideal dataset for our study purpose.

While this dataset provides a unique resource, it also imposes specific constraints on the analysis of this study. The longitudinal, observational nature of the GRLS means that laboratory data were collected annually without a fixed schedule relative to cancer diagnosis. Consequently, the analytical dataset inherently includes a mixture of pre-diagnostic, peri-diagnostic, and post-diagnostic visits, the latter of which may be confounded by treatment effects. Furthermore, the distribution of cancer cases across many different tumor types within the available data necessitated a multi-cancer approach for this initial benchmark, which may bias the model toward generic rather than cancer-specific signals. These inherent characteristics of the dataset define the realistic conditions under which we test the feasibility of using routine lab data for early cancer detection.

Given these challenges, this study was designed not to develop a clinical tool, but to establish a much-needed, rigorous performance benchmark. The goal is to quantify the maximal predictive performance achievable using only routine laboratory data from a large, longitudinal canine cohort. By systematically evaluating a wide range of ML architectures and leveraging Explainable AI (XAI) [24], this study provides a transparent assessment of the inherent limitations of this data modality when used in isolation and a methodological guide for future research.

## 2. Methods

### 2.1 Data Source

Data for this study were sourced from the MAF GRLS Data Commons Portal. The GRLS aims to identify genetic, environmental, and lifestyle risk factors for cancer and other common diseases by following dogs enrolled between June 2012 and April 2015 from a young age (6 months to 2 years at enrollment) for their entire lives [14, 15].

GRLS data collection is performed annually through three components: an owner questionnaire, a veterinarian examination and questionnaire, and the collection of biospecimens. During the annual veterinary visits, a full physical examination is performed and core samples, including whole blood and serum, are collected. These samples are sent to a veterinary diagnostic reference laboratory for CBC and Chem profile, and the results are added to the GRLS database. CBC is run on a Siemens Advia, and chemistry is run on Beckman Coulter AU-series for the whole duration of the GRLS study. CBC and chemistry were run in one lab from 2012 to Dec. 2020, and another lab from 2021 to April 2023. During this change, a bridging study where samples from 50 dogs were run at both labs [15], confirmed that the change did not impact the results.

For the specific purpose of cancer ascertainment, in addition to the annual follow-up, owners and veterinarians are instructed to contact the study in the event of a suspected or confirmed malignancy diagnosis or death. When a malignancy is suspected, the study requests that veterinarians submit biopsy samples for histology. All cancer diagnoses within the GRLS are assigned one of three tiers of confidence [15]. Tier 1 is a definitive diagnosis that is microscopically confirmed by a board-certified pathologist via histology or cytology, Tier 2 is a presumptive diagnosis based on direct visualization, imaging, or an in-house cytology reading, and Tier 3 is a presumptive diagnosis based on clinical suspicion alone.

All diagnostic tiers were captured in the current study to maximize case identification. For deceased dogs, full medical records were obtained, and necropsies were performed at the owner's discretion to confirm the cause of death [15]. The GRLS study does not actively "confirm" dogs to be non-diseased in a proactive, screening sense. The non-diseased (control) status is defined by absence of a reported diagnosis.

## 2.2 Data Curation and Case Definition

A multi-step data curation pipeline was implemented to construct the final analytical dataset for model training. The process is outlined below and summarized in Figure 1.

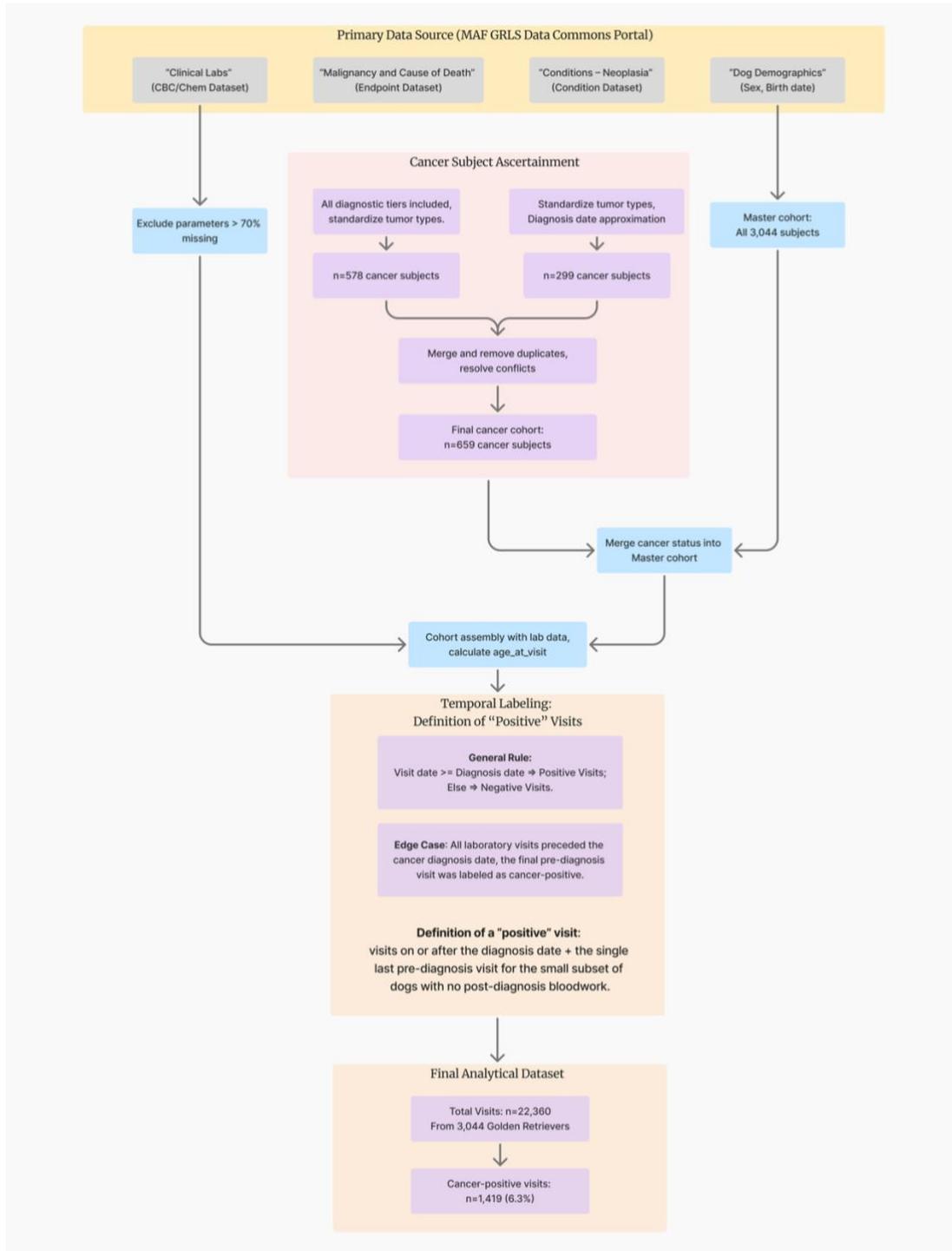

**Figure 1. Data Curation Pipeline.** This flowchart illustrates the multi-step process of constructing the final analytical dataset from the Golden Retriever Lifetime Study (GRLS) data. The pipeline includes data extraction from multiple sources, cancer subject ascertainment, cohort assembly, and temporal labeling of veterinary visits.

### 2.2.1 Raw Data and Initial Processing

Raw data were extracted from the MAF Data Commons Portal. The primary sources included:

- "Dog Demographics": For subject sex and birth date.
- "Clinical Labs" (CBC/Chem Dataset): For CBC (297,678 visits from 3,044 dogs) and Chem (547,519 visits from 3,043 dogs) profiles.
- Cancer Diagnoses (two sources were used to ensure comprehensive case ascertainment)
    - "Malignancy and Cause of Death" (Endpoint Dataset): Considered the primary source, containing diagnoses from medical records and diagnostics.
    - "Conditions – Neoplasia" (Condition Dataset): A secondary source derived from annual and ad-hoc veterinarian questionnaires.

### 2.2.2 Cancer Subject Ascertainment

To define the diseased cohort (cancer subjects), we integrated the two cancer data sources.

Endpoint Dataset Processing: All malignancy diagnoses, regardless of diagnostic confidence tier (Tier 1: 73.8%; Tier 2: 16.9%; Tier 3: 9.3%, 12 entries have missing values for tier information), were included to maximize sensitivity. A manual review of the tracked_conditions column identified 76 distinct cancer types (see Supplementary Material S1.1), which were used to flag subjects in the Endpoint Dataset as cancer patients. This yielded 578 cancer subjects.

Condition Dataset Processing: Subjects were flagged as cancer patients if the "any" column indicated a diagnosis (1 = at least one cancer condition). This identified 299 subjects, and their unique cancer types (34 distinct types) were compiled (see Supplementary Material S1.2).

Diagnosis Date Approximation for Condition Dataset: Since precise diagnosis dates were unavailable in the Condition dataset, we derived approximate diagnosis dates by combining the 'record_date' (questionnaire submission date) and 'to_date' columns. The 'to_date' column was used to filter for newly diagnosed conditions (to_date = 0, indicating conditions diagnosed only in that study year), ensuring we captured incident cases. From the filtered data, we extracted the earliest date from each 'record_date' string, consolidated tumor-specific columns, and assigned the earliest record_date for each subject-tumor combination as the diagnosis_date, providing the most accurate approximation of initial diagnosis timing.

### 2.2.3 Dataset Merging and Final Cancer Cohort

The two cancer datasets were merged to create a unified cancer cohort. For subjects with multiple cancer diagnoses, only the entry with the earliest diagnosis_date was retained to record the first cancer occurrence. A full outer join was performed on subject_id to include all unique cancer subjects from both sources.

For conflicting information, the Endpoint dataset was prioritized: its diagnosis_date was used when available, and its tumor type designation was preferred. Tumor type names were standardized to primary categories (e.g., "Hemangiosarcoma – cardiac" to "hemangiosarcoma"; see Supplementary Material S1.3 for Tumor Type Standardization Protocol and Final Tumor Categories).

This process yielded a final cancer cohort of 659 unique subjects. The top 10 tumor types were: hemangiosarcoma (25.55%), mast cell tumor (17.43%), lymphoma (15.21%), soft tissue sarcoma (8.86%), histiocytic sarcoma (4.73%), melanoma (2.81%), CNS tumor (2.66%), eye tumor (2.51%), osteosarcoma (1.77%), and leukemia (1.77%). The diagnosis date for 578 subjects (87.7%) was sourced from the higher-quality Endpoint dataset, mitigating potential inaccuracies from the date approximation.

### 2.2.4 Cohort Assembly and Temporal Labeling

The final cancer cohort was merged with the 'Dog Demographics' dataset to create a master cohort of all 3,044 dogs. Cancer subjects were flagged (cancer_status = 1), and their tumor type and diagnosis date were appended. This was then merged with the cleaned CBC and Chem data from the 'Clinical Labs' dataset. Lab parameters with >70% missing values were excluded. Age at each visit was calculated as visit_date minus birth_date.

To prevent data leakage in the predictive model, a temporal adjustment protocol was applied to label veterinary visits correctly. A visit was labeled as positive (Tumor = 1) only if the visit_date occurred on or after the patient's cancer diagnosis_date. All visits prior to the diagnosis date were labeled as negative (Tumor = 0). For 272 canine subjects (41.4% of all cancer subjects), all laboratory visits preceded the cancer diagnosis date. In these cases, only the final pre-diagnosis visit was labeled as cancer-positive to preserve these minority-class examples (19% of cancer-positive visits) while maintaining temporal accuracy.

Consequently, the final definition of a "positive visit" included: 1) all visits on or after the diagnosis date, and 2) the single last pre-diagnosis visit for subjects with no post-diagnosis laboratory data.

Key implications of this labeling strategy were:

- The term "last pre-diagnosis visit" was not defined by a fixed duration but simply as the last available lab visit before the recorded diagnosis date due to the cadence of visits (annually).
- For subjects with multiple primary cancers, only the first cancer diagnosis was used to define the diagnosis_date; all subsequent visits were labeled as positive from this date onward.
- The model was trained without discriminating for treatment status, meaning it was exposed to lab results from both treated and untreated periods following diagnosis.

### 2.2.5 Final Analytical Dataset

The final curated dataset comprised 22,460 unique veterinary visits from 3,044 dogs, including 659 cancer subjects. It contains demographic information (sex, age at visit), CBC and Chem parameters, and cancer status (with diagnosis date and tumor type for positive cases) for use as features in machine learning. The severe class imbalance at the visit level (Table 1.1) is the central statistical challenge of this study. Detailed analysis of missing values and feature correlations is provided in the Supplementary Material S2.2.

The demographic and clinical characteristics of the final working dataset are summarized in Table 1.1, which highlights the severe class imbalance at the visit level (6.3% cancer-positive), representing the central statistical challenge for model development.

The relationship between age and cancer-positive visit rate is detailed in Table 1.2, demonstrating a strong, non-linear increase in cancer risk with advancing age, which is a major biological and clinical characteristic of the dataset that the model must contend with.

**Table 1.1. Demographic and Clinical Overview of the Final Working Dataset.** Data were collected from annual veterinary visits between June 2012 and April 2023. Note: due to embrago period of GRLS dataset, not all results were published on the MAF GRLS data commons at the time the study was conducted.

| Key Metric | Value |
| --- | --- |
| **Data Collection Period** | June 2012 - April 2023 |
| **Total Subjects (n)** | 3,044 |
| **Total Visits (n)** | 22,460 |
| **Age Characteristics** | Mean Age ± Standard Deviation (Years) |
| **Mean age of cancer subjects at first diagnosis (range)** | 7.3 ± 1.9 (1.1 – 10.3) |
| **Mean age of non-cancer subjects at last visit (range)** | 7.6 ± 2.8 (0.5 - 12.9) |
| **Overall mean age range across all visits (range)** | 4.9 ± 2.6 (0.4 - 12.9) |
| **Sex Distribution** | |
| Male (n, %) | 1,540 (50.6%) |
| Female (n, %) | 1,504 (49.4%) |
| **Cancer Status Distribution** | |
| Subjects diagnosed with cancer | 659 (21.6%) |
| Subjects without a cancer diagnosis | 2,385 (78.4%) |

| Visit-Level Cancer Status | |
|---|---|
| Visits labeled as cancer-positive | 1,419 (6.3%) |
| Visits labeled as cancer-negative | 21,115 (93.7%) |

*Note: "Mean age of cancer subjects at first diagnosis" represents the age at initial cancer diagnosis. "Mean age of non-cancer subjects at last visit" indicates the duration of follow-up for control subjects. "Overall age range across all visits" represents the average age across all 22,460 visits. Age ranges represent minimum to maximum values.*

**Table 1.2. Distribution of Cancer-Positive Visits by Age at Visit.**

| Age (Years) | Total Visits (n, %) | Cancer-Positive Visits (n) | Cancer-Positive Visit Rate (%) |
|---|---|---|---|
| 0-2 | **3,927 (17.48%)** | 8 | 0.20 |
| 2-4 | **5,507 (24.52%)** | 67 | 1.22 |
| 4-6 | **4,988 (22.21%)** | 208 | 4.17 |
| 6-8 | **4,576 (20.37%)** | 505 | 11.04 |
| 8+ | **3,462 (15.41%)** | 631 | 18.23 |
| **Total** | **22,460 (100%)** | 1,419 | 6.32 |

*Note: This table shows the distribution of visits labeled as "cancer-positive" based on the age of the dog at the time of the bloodwork. A visit was labeled as cancer-positive if it occurred on or after the date of the first cancer diagnosis, or if it was the single last pre-diagnosis visit for dogs with no post-diagnosis laboratory data, as defined in Section 2.2.4. The "Cancer-Positive Visit Rate" is the proportion of visits within each age bracket that were labeled as positive.*

**2.3 Feature Engineering and Imputation**

The initial feature set included CBC and biochemistry parameters, age at visit, and sex (see Supplementary Material S2.1 for Complete List of Features).

Two composite ratios were engineered based on their established association with systemic inflammation in a variety of canine diseases: the Neutrophil-to-Lymphocyte Ratio (NLR) and the Platelet-to-Lymphocyte Ratio (PLR). The NLR has been demonstrated to be elevated in non-neoplastic inflammatory states such as inflammatory bowel disease [25], meningoencephalitis of unknown etiology [26], and acute congestive heart failure [27], and has been identified as a potential marker in canine leishmaniasis [28]. The PLR is a recognized marker in inflammatory conditions like acute pancreatitis [29]. Beyond their role in general inflammation, these ratios have been investigated for their prognostic value in specific canine cancers. The NLR has been shown to be prognostic in mammary tumors [30], lymphoma in small-breed dogs [31], and osteosarcoma [32]. Their inclusion was therefore justified as potential indicators of the systemic inflammatory response often associated with malignancy and an overall poorer prognosis.

Missing data were imputed using Multivariate Imputation by Chained Equations (MICE), which preserves correlations between variables by iteratively modeling each feature as a function of other features in the dataset [33]. This approach generates more plausible imputations than simple methods by leveraging the underlying multivariate structure of the data. All numerical data were then adjusted to a common scale (using RobustScaler) to ensure no single test result unfairly dominated the analysis due to its naturally larger numbers.

**2.4 Model Development and Validation Protocol**

A rigorous, comparative framework was used to identify the optimal modeling pipeline.

**Data Splitting**: The dataset was partitioned into training (60%), validation (20%), and test (20%) sets using GroupShuffleSplit, with subject_id as the grouping variable. The training set is used to build the models, the validation set is used to compare and select the best-performing model during development, and the held-out test set provides a final, unbiased estimate of how the chosen model would perform on new, unseen patients. This patient-level split prevents data leakage by ensuring all visits from a single dog are confined to one data partition. This is critical because if visits from the same dog were spread across different sets, the model could effectively "memorize" individual patients' patterns, invalidating the test as a true assessment of generalization. Visual inspection confirmed that the distribution of the cancer-positive visits was similar across all three splits, ensuring that the evaluation sets were representative of the overall data distribution.

**Comparative Framework:** A systematic, comparative approach was used to identify the best-performing combination of a machine learning algorithm, a feature set and a data balancing technique.

- **Base Models:** A diverse set of six base models was evaluated to cover different algorithmic families: Logistic Regression (LR), Random Forest Classifier (RF), XGBoost Classifier, LightGBM Classifier (two powerful gradient boosting methods), a Multi-Layer Perceptron (a neural network architecture), and Naïve Bayes.
- **Feature Selection**: Three distinct feature sets were generated for model comparison. For the two automated sets, an optimal feature count (k) was determined using Recursive Feature Elimination (RFE) with Cross-Validation. This process used a LightGBM classifier over 5 folds to maximize the ROC-AUC score. This optimal k was then used to create a univariate set with SelectKBest and a multivariate set with RFE using a RF estimator. These were compared against a third, manually curated set of 15 biomarkers, selected for their established clinical relevance to common paraneoplastic syndromes like anemia, thrombocytopenia, and hypercalcemia (see Supplementary Material for the full list).
- **Data Balancing Techniques**: Six resampling techniques (SMOTE, ADASYN, RandomOverSampler, RandomUnderSampler, SMOTETomek, SMOTEEnn) were compared against a baseline using class weighting within the model (LR and RF) or with no resampling applied.

**Hyperparameter Tuning and Selection:** Each of the 126 (6x3x7) pipelines underwent 5-fold cross-validated GridSearchCV on the training set, optimizing for the Matthews Correlation Coefficient (MCC) [34]. Hyperparameters (see examples in Table 3.2) are the configuration settings of a model that are not learned directly from the data. Hyperparameter tuning is the process of finding the optimal combination of these settings. GridSearchCV is an exhaustive method that automatically tests a predefined set of hyperparameter values and uses cross-validation to identify the best combination. The MCC is a single metric that considers all four categories of the confusion matrix (true positives, false positives, true negatives, false negatives). It is regarded as a balanced measure that can be used even when the classes are of very different sizes, making it particularly suitable for optimizing model performance on imbalanced datasets like the one used in this study. The single best pipeline was selected based on its performance on the validation set, primarily assessed by MCC and Area Under the Receiver Operating Characteristic Curve (AUROC or AUC or ROC-AUC score).

Analyses were performed in Python (3.8.8) using scikit-learn (1.3.2), XGBoost (2.1.4), LightGBM (4.6.0), and SHAP (0.44.1).

### 2.5 Final Model Evaluation and Interpretation

The selected pipeline was retrained on the combined training and validation sets and evaluated once on the test set. Performance was assessed using MCC, ROC-AUC (with 95% CI), precision, recall, F1-score, and Positive/Negative Predictive Values (PPV/NPV). Both ROC and Precision-Recall (PR) curves were generated. Although it was widely considered that ROC-AUC is prone to inflation when it comes to imbalanced datasets, a recent study has shown that the ROC curve is indeed robust to class imbalance through simulated and real-world data while the PR curve is highly sensitive to class imbalance [35]. They encourage the adoption of ROC-AUC in such cases for fairer comparisons of models across datasets with different imbalances and furthering the understanding of the relationship between the ROC and PR spaces.

Model interpretation was performed using SHAP (SHapley Additive exPlanations) [36]. SHAP is a method based on cooperative game theory that quantifies the contribution of each input feature (e.g., a lab value) to the final prediction for an individual instance. It provides both global insight into overall feature importance and local explanations for single predictions.

## 3. Results

### 3.1 Final Cohort Characteristics

Following the data curation and temporal adjustment, the final analytical dataset consisted of 22,460 clinical lab visits from 3,044 Golden Retrievers. As detailed in Table 1, the key characteristic of this dataset is the low prevalence of the target class, with cancer-positive visits accounting for only 6.3% of the total observations. This severe class imbalance is the defining challenge for the classification task.

### 3.2 Performance of the Final Optimized Model on the Test Set

The primary finding of this study is the gap between the model's ability to rank patients by risk and its ability to accurately classify them. Table 3.1 summarizes the top 10 performing model combinations. As illustrated in the table, the Logistic Regression model using class weighting on the RFE feature set achieved the highest MCC and AUC on the validation set and was therefore selected for the final evaluation. This result provides powerful evidence that the performance limitation is inherent to the data itself, not a failure to select an appropriate analytical method. Table 3.2 lists the optimized hyperparameters for the top 10 models identified during the selection process. Please find the comprehensive hyperparameter search space for all ML models in the Supplentary Material S3.1.

**Table 3.1: Comparative Performance of Top 10 Model-Sampler Combinations on the Validation Set, Ranked by their MCC score.**

| Base Model | Data Balancer | Feature Set | MCC | AUC |
|---|---|---|---|---|
| **LogisticRegression** | None | rfe | 0.2800 | 0.8292 |
| **LGBM** | RandomUnderSampler | manual | 0.2756 | 0.8237 |
| **LGBM** | SMOTEEnn | manual | 0.2728 | 0.8108 |
| **LogisticRegression** | None | univariate | 0.2672 | 0.8246 |
| **LogisticRegression** | None | manual | 0.2653 | 0.8242 |
| **XGB** | RandomUnderSampler | manual | 0.2645 | 0.8133 |
| **RandomForest** | RandomUnderSampler | univariate | 0.2603 | 0.7973 |
| **LGBM** | RandomUnderSampler | rfe | 0.2592 | 0.7983 |
| **MLP** | RandomUnderSampler | manual | 0.2579 | 0.8039 |
| **NaiveBayes** | ADASYN | manual | 0.2552 | 0.8010 |

**Table 3.2: Optimized Hyperparameters for Top 10 Models.**

| Base Model | Optimized Hyperparameters |
|---|---|
| **LogisticRegression** | C: 0.1 |
| **LGBM** | learning_rate: 0.01, max_depth: 3, n_estimators: 300 |
| **LGBM** | learning_rate: 0.01, max_depth: 9, n_estimators: 200 |
| **LogisticRegression** | C: 0.1 |
| **LogisticRegression** | C: 0.1 |
| **XGB** | learning_rate: 0.01, max_depth: 6, n_estimators: 300, subsample: 0.9 |
| **RandomForest** | max_depth: None, min_samples_leaf: 1, min_samples_split: 2, n_estimators: 50 |
| **LGBM** | learning_rate: 0.01, max_depth: 9, n_estimators: 200 |
| **MLP** | alpha: 0.1, hidden_layer_sizes: (50,), learning_rate: 'constant' |
| **NaiveBayes** | None |

**Table 3.3: Final Model Performance on the Test Set.**

|  | MCC | AUC(95% CI: 0.793-0.836) | PPV | NPV | Recall | Specificity | Accuracy |
|---|---|---|---|---|---|---|---|
| **Score** | 0.2537 | 0.8153 | 0.1464 | 0.9819 | 0.7917 | 0.7102 | 0.7151 |

The final optimized Logistic Regression model achieved a strong and statistically significant AUC of 0.815 on the test set. However, this did not translate into effective classification performance, as reflected by the clinically unacceptable F1-score of 0.25 and PPV of 0.15. These conflicting results are visually summarized in Figure 2. The ROC curve (Figure 2A) demonstrates good discriminatory ability, sitting well above the baseline of random chance. In contrast, the PR curve (Figure 2B), which is more sensitive to class imbalance, illustrates the model's practical failure. To achieve any meaningful recall, precision drops to a level that would generate an overwhelming number of false positives in a clinical setting. The complete performance metrics on the test set are presented in Table 3.3.

The strong AUC indicates that the model has learned a genuine signal within the data; a randomly selected cancer-positive visit has an 81.5% chance of being assigned a higher risk score by the model than a randomly selected cancer-negative visit. The high NPV of 0.98 suggests that the model is good at ruling out disease. However, the extremely low Precision (PPV) of 0.15 is the model's major failure. This means that of all the visits the model flags as "high-risk," only 15% are actual cancer cases, while the remaining 85% are false positives. The low F1-score of 0.25, which is the harmonic mean of precision and recall, encapsulates this poor balance and confirms the model's lack of clinical utility for positive case identification.

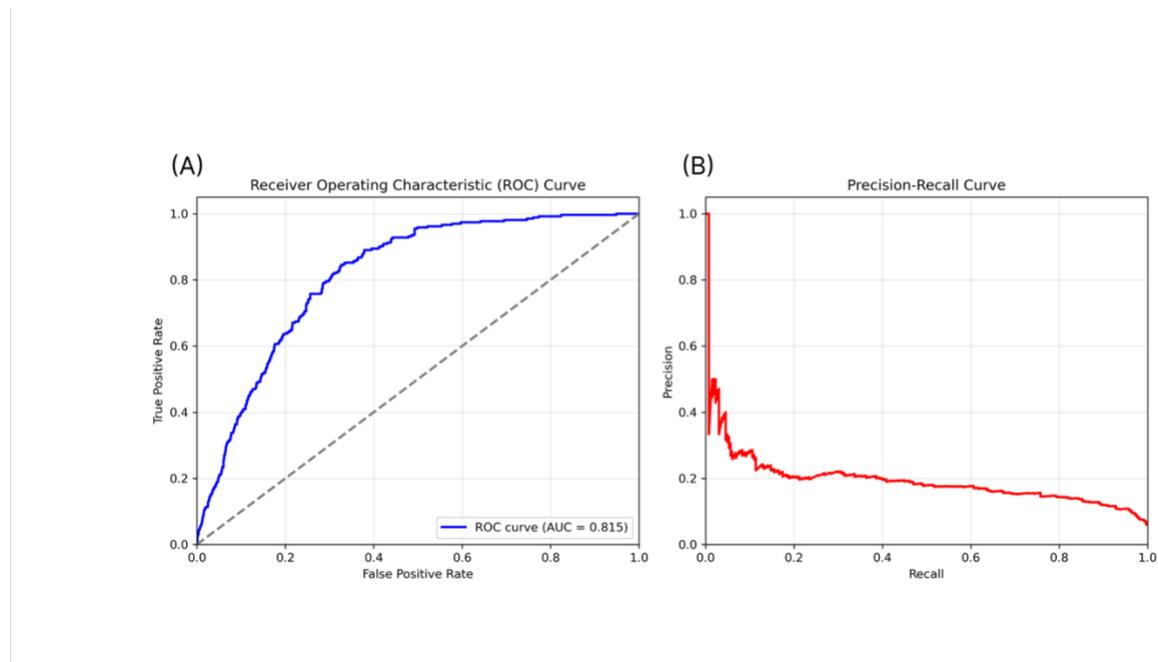

**Figure 2: Final Model Classification Performance on the Test Set**. (A) The Receiver Operating Characteristic (ROC) curve plots the true positive rate against the false positive rate. An Area Under the Curve (AUC) of 0.815 demonstrates a moderate ability to discriminate between cancer-positive and cancer-negative visits, performing significantly better than random chance (dashed line). (B) The Precision-Recall (PR) curve plots precision against recall and is more informative

for imbalanced datasets. The curve sits only slightly above the no-skill baseline (equal to the cancer prevalence of 6.3%), visually demonstrating the model's clinical failure. To achieve meaningful recall (sensitivity), precision drops to an unacceptably low level, which would result in a high number of false positives.

### 3.3 Drivers of Model Prediction: An Explainable AI Analysis

The SHAP analysis provides a clear explanation for the model's performance paradox. As shown in Figure 3, predictions were overwhelmingly driven by biologically plausible but highly non-specific features.

Patient age was the single most powerful predictor, followed by features related to anemia (e.g., lower hemoglobin) and inflammation (e.g., higher band neutrophils, higher NLR). This reveals that the model effectively learned to identify older dogs with signs of chronic disease rather than a specific signature of cancer.

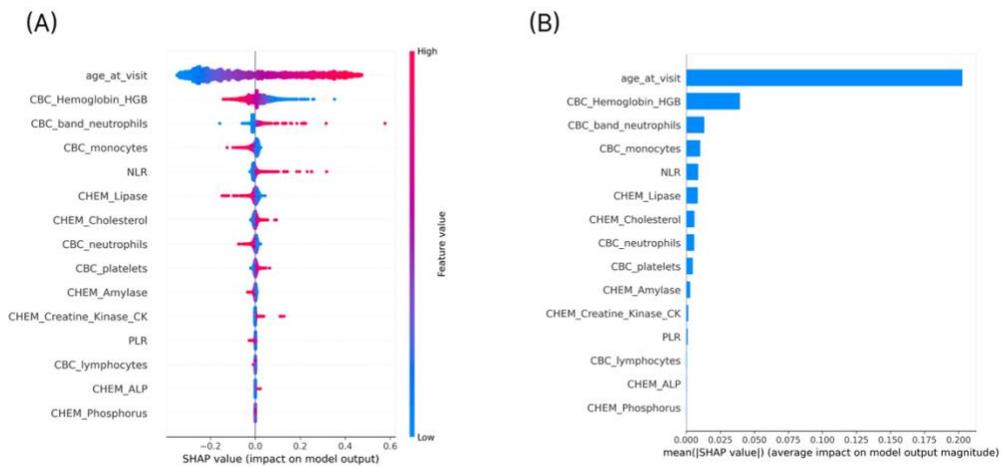

**Figure 3: SHAP Analysis of Feature Importance.** (A) The Global Feature Importance plot ranks features by their mean absolute SHAP value, or their average impact on the model's prediction magnitude. (B) The SHAP summary plot shows the impact of each feature's value on the model's prediction. Each dot represents a single visit; red indicates a high feature value and blue indicates a low feature value.

## 4. Discussion

The results present a common inconsistency in machine learning applications on imbalanced datasets: a fairly good ROC curve (Figure 2A) alongside a poor PR curve (Figure 2B). It indicates that the model has identified a genuine, albeit weak, signal within the multivariate data. This modest performance should be interpreted within the context of a dataset designed to reflect real-world conditions; for the 41.4% of cancer subjects with no post-diagnosis lab work, only their final pre-diagnosis visit was labeled as positive, simulating the recognized diagnostic gap in veterinary care [4]. While this confirms a statistical association among bloodwork, age, sex, and cancer status, enabling effective patient ranking, the severe overlap in risk score distributions between the cancer and non-cancer cohorts precludes the selection of a classification threshold that can cleanly separate these groups. This results in a low F1-score and highlights the clinically unacceptable trade-off between precision and recall.

The model's high NPV of 0.98 might suggest a potential clinical application as a "rule-out" tool, wherein patients with a negative prediction would be cleared without further investigation, potentially reducing a clinician's caseload [37]. This application is, however, untenable upon closer examination of the model's recall. A recall of 0.79 on the test set indicates that the model fails to identify 21% of dogs with cancer. The clinical consequences of this false-negative rate, delaying or missing the diagnosis in nearly one-quarter of affected patients, would be severe, rendering the model unsafe for this purpose.

SHAP analysis revealed that the model's most influential predictors, while biologically plausible, are non-specific. Age was the most powerful predictor, which is expected as cancer is predominantly a disease of aging. However, a model heavily reliant on age risks functioning more as a proxy for age-related morbidity, similar to an "old dog detector", than as a specific cancer biomarker. Beyond age, the model identified patterns consistent with anemia of chronic disease (e.g., decreased hemoglobin) and systemic inflammation (e.g., increased band neutrophils, NLR). The primary limitation is the model's inability to distinguish cancer-induced changes from those caused by other common geriatric conditions, such as chronic kidney disease, inflammatory bowel disease, or immune-mediated polyarthritis. This lack of specificity is the driver of the high false-positive rate.

Perhaps the most significant limitation is time-varying confounding by treatment [38-40], a direct consequence of the study design which included post-diagnosis visits to benchmark real-world data utility. This phenomenon occurs when a treatment administered after diagnosis affects the same predictors the model uses for classification. The observational nature of the dataset captures this process. For instance, a dog with lymphoma treated with a CHOP protocol, a common multi-agent chemotherapy regimen comprising Cyclophosphamide, Hydroxydaunorubicin (doxorubicin), Oncovin (vincristine), and Prednisonestudy, will develop a stress leukogram and potential hyperglycemia. Similarly, a dog with a mast cell tumor treated with a tyrosine kinase inhibitor like toceranib phosphate (Palladia®) may develop gastrointestinal toxicity leading to hypoalbuminemia. As our methodological design included post-diagnosis visits, the model is likely learning to associate these iatrogenic effects of treatment with the cancer label, not the underlying signature of the disease itself.

This reveals a flaw in the model's causal inference. The model has not learned the desired causal pathway (Cancer → Blood Changes) but rather a confounded one (Cancer → Diagnosis → Treatment → Blood Changes). Consequently, the model's predictive power may be significantly reliant on signals that only emerge after treatment has begun, rendering it ineffective for its intended use in a pre-diagnosis, treatment-naive screening population. Addressing this confounder requires advanced causal inference methods, such as marginal structural models or Single World Intervention Graphs [41, 42], which represent a necessary direction for future research.

It is important to frame the study's conclusions within the context of its strengths and limitations. The primary strengths include the use of a large, longitudinal cohort, the methodologically rigorous patient-level data splitting to prevent leakage, the systematic and transparent comparison of multiple models and balancing techniques, and the use of advanced XAI for deep model interpretation. These strengths ensure that the negative findings are robust and not merely an artifact of a suboptimal analytical approach.

However, several key limitations must be acknowledged. First, as detailed in the Methods, the inclusion of post-diagnosis visits without accounting for treatment status is a major limitation that confounds the model's learned signals with the effects of therapy. This design choice was made to benchmark performance under real-world data conditions, but it fundamentally limits the model's utility for presymptomatic screening. Second, the multi-cancer approach taken here represents a significant constraint. This was not an intentional design choice but a necessary concession to data limitations; because of the embargoed period of GRLS dataset, at the time the study was conducted, the accessed number of cases for any single cancer type was too low to train a robust machine learning model. By grouping dozens of histologically distinct malignancies (e.g., hemangiosarcoma, lymphoma, mast cell tumors), we forced the model to find a common signal across cancers with vastly different pathophysiologies. This inevitably biased it toward detecting generic markers of systemic illness, such as inflammation and anemia, rather than specific oncologic signals, which inherently limited the model's specificity and contributed to the high false-positive rate. A single-cancer approach would undoubtedly be more effective, and it should be the next step for future research as more case data for specific malignancies accrues in the cohort. Third, the study population was restricted to Golden Retrievers, a breed with a known genetic predisposition to specific cancers like hemangiosarcoma and lymphoma [43], which limits the generalizability of our findings. Future work should seek to validate these results in breeds with different cancer predispositions (e.g., Boxers, Scottish Terriers) [44, 45]. Finally, by design, the model was restricted to laboratory data, age, and sex. It lacked access to crucial clinical context such as physical exam findings, co-morbidities, or owner-reported signs. While this was intended to test the feasibility of a bloodwork-only screening tool, the results underscore the importance of these unmeasured variables.

The clinical implications of this study are direct: the model developed herein, and likely any model based solely on this data modality, is unsuitable for clinical deployment. Its low PPV would generate a high number of false positives, leading to significant potential for harm through unnecessary diagnostic procedures, owner anxiety, and misallocation of veterinary resources.

The primary value of this work lies in establishing a realistic, data-driven performance benchmark for cancer screening using only routine hematological data. It demonstrates with methodological rigor that the path forward is not through more complex algorithms but through a fundamental shift towards multi-modal data integration. This aligns with a growing body of literature in veterinary oncology that demonstrates the power of integrating diverse data types, such as medical records, imaging data (radiomics, digital pathology), molecular diagnostics (e.g., flow cytometry, PCR), and other modalities of data (e.g., drug sensitivity, X-ray diffraction, themodymatic data) [46-52]. Future models must be designed to incorporate the same rich, multi-faceted information a skilled clinician uses, including signalment, clinical history, and physical exam findings. While integrating advanced diagnostics like genomics or liquid biopsies promises greater specificity, researchers must remain mindful that this may compromise the accessibility of resulting models for general practices.

Ultimately, this study underscores the need for large-scale, multi-center collaborations to build and validate robust, generalizable models from datasets that are both multi-modal and representative of diverse patient populations. Independent validation in large, external cohorts is an essential prerequisite before any such tool can be considered for clinical use.

## 5. Conclusion

This study provides a rigorous feasibility assessment and establishes a performance benchmark for using routine canine laboratory data for cancer detection. The findings demonstrate that while such data contain a statistically detectable signal associated with malignancy, it is ultimately insufficient for developing a clinically reliable classification tool. The severe overlap between the hematological signatures of cancer, aging, and other non-neoplastic inflammatory conditions, compounded by the critical challenge of treatment-related confounding in observational data and the inherent limitations of a multi-cancer approach, results in a model with clinically unacceptable performance.

The conclusion is not that machine learning has no role in veterinary oncology, but rather that its application must be more sophisticated and data-aware. This work establishes a crucial benchmark, illustrating the performance ceiling of a single-modality approach when applied to real-world observational data. The future of AI in veterinary cancer diagnostics lies not in seeking one "gold coin" within this limited data stream, but in the integration of multi-modal data to create a holistic patient representation that more closely emulates the diagnostic reasoning of an expert clinician.

**List of Abbreviations:**

Complete Blood Count (CBC)

Serum Biochemistry panels (Chem)

Machine Learning (ML)

Morris Animal Foundation (MAF)

Golden Retriever Lifetime Study (GRLS)

Explainable AI (XAI)

Neutrophil-to-Lymphocyte Ratio (NLR)

Platelet-to-Lymphocyte Ratio (PLR)

Multivariate imputation by chained equations (MICE)

Logistic Regression (LR)

Random Forest Classifier (RF)

Recursive Feature Elimination (RFE)

Matthews Correlation Coefficient (MCC)

Area Under the Receiver Operating Characteristic Curve (AUROC or AUC or ROC-AUC score)

Positive/Negative Predictive Values (PPV/NPV)

Precision-Recall Curve (PR Curve)

SHapley Additive exPlanations (SHAP)

**Acknowledgement:**

This work was made possible by the Morris Animal Foundation, its staff, and the generous contributions of the Golden Retriever Lifetime Study community.

**Data avaiability:**

The dataset for this study is available upon request from the Morris Animal Foundation's Data Commons: https://datacommons.morrisanimalfoundation.org/. The source code is available at:

https://github.com/Shumin-li-mcit/canine-cancer-lab-data-benchmark.

# Supplemental material

## S1. Cancer Subject Ascertainment

### S1.1: Raw Tumor Types from the Endpoint Dataset

This list contains the keywords and phrases used to identify cancer-related entries/tumor types in the 'tracked_conditions' column of the "Malignancy and Cause of Death" (Endpoint) dataset. A total of 79 distinct terms were identified, listed below in alphabetical order.

"Acanthomatous ameloblastoma, Adenocarcinoma - apocrine gland anal sac, Adenocarcinoma - exocrine pancreatic, Adenocarcinoma - mammary, Adenocarcinoma - other/not specified, Anaplastic sarcoma, Apocrine gland ductal adenoma, Brain tumor, Carcinoma - basal cell, Carcinoma - basosquamous cell, Carcinoma - choroid plexus, Carcinoma - gastric, Carcinoma - hepatocellular, Carcinoma - mammary, Carcinoma - nasal, Carcinoma - neuroendocrine, Carcinoma - other/not specified, Carcinoma - ovarian, Carcinoma - pulmonary, Carcinoma - squamous cell, Carcinoma - thyroid, Carcinoma - transitional cell, Carcinosarcoma - thyroid, Chondrosarcoma, CNS tumor, Cutaneous melanoma, Gastrointestinal stromal tumor, Hemangiosarcoma - cardiac, Hemangiosarcoma - cutaneous, Hemangiosarcoma - other/not specified, Hemangiosarcoma - splenic, Hemangiosarcoma - visceral, Histiocytic sarcoma, Leiomyosarcoma, Leukemia, Liposarcoma, Liver tumor, Lymphoma - cutaneous, Lymphoma - gastrointestinal, Lymphoma - multicentric, Lymphoma - other/not specified, Malignant melanoma, Malignant pilomatricoma, Malignant trichoepithelioma, Mast cell tumor - cutaneous, Mast cell tumor - other/not specified, Mast cell tumor - subcutaneous, Meibomian gland epithelioma, Meningioma, Metastatic sarcoma, Multiple myeloma, Myelodysplastic syndrome, Nasal sarcoma, Nasal tumor, Nephroblastoma, Oral melanoma, Osteosarcoma - appendicular, Osteosarcoma - axial, Osteosarcoma - other/unspecified, Pituitary adenoma, Plasma cell tumor, Rhabdomyosarcoma, Round cell tumor, Sarcoma, Soft tissue sarcoma - fibrosarcoma, Soft tissue sarcoma - giant cell tumor, Soft tissue sarcoma - keloidal, Soft tissue sarcoma - myxosarcoma, Soft tissue sarcoma - other/not specified, Soft tissue sarcoma - perivascular wall tumor, Soft tissue sarcoma - peripheral nerve sheath tumor, Soft tissue sarcoma - spindle cell sarcoma, Soft tissue sarcoma - synovial cell sarcoma, Soft tissue sarcoma - undifferentiated sarcoma, Spleen tumor, Testicular tumor, Thymoma, Undifferentiated malignant neoplasm, Unknown neoplasia."

### S1.2: Raw Tumor Types from the Condition Dataset

This list contains the raw, unstandardized tumor types as originally reported in the "Conditions – Neoplasia" (Condition) dataset. A total of 34 distinct terms were identified, listed below in alphabetical order.

"adrenal_tumor, basal_cell_tumor, bile_duct_tumor, bladder_tumor, brain_spinal_cord_tumor, breast_or_mammary_tumor, eye_tumor, hair_matrix_tumor, heart_tumor, hemangiosarcoma, histiocytic_sarcoma, kidney_tumor, leukemia, liver_tumor, lung_tumor, lymphoma, mast_cell_tumor, melanoma, multiple_myeloma, nasal_tumor, osteosarcoma, pancreatic_tumor, perianal_adenoma, pituitary_tumor, plasma_cell_tumor, plasmacytoma, prostate_tumor,

soft_tissue_sarcoma, splenic_tumor, squamous_cell_carcinoma, stomach_intestinal_tumor, testicular_tumor, thymoma, thyroid_tumor."

### S1.3: Tumor Type Standardization Protocol and Final Tumor Categories

To create a consistent set of tumor categories for analysis, the specific tumor type names from both source datasets were standardized into primary categories. The main consolidation rules were: 1) Variants of the same cancer type (e.g., 'Lymphoma - multicentric', 'Lymphoma - cutaneous') were consolidated into a single primary category (e.g., 'lymphoma'); 2) Carcinomas and adenocarcinomas were grouped by organ system; 3) The resulting standardized types were converted to a consistent snake_case format. This process resulted in 45 distinct standardized tumor categories in the final analytical dataset, listed below in alphabetical order.

"adenocarcinoma_anal_sac, adenocarcinoma_mammary, adenocarcinoma_other, basal_cell_tumor, benign_skin_tumor, carcinoma_gastric, carcinoma_hepatic, carcinoma_mammary, carcinoma_nasal, carcinoma_neuroendocrine, carcinoma_other, carcinoma_ovarian, carcinoma_pulmonary, carcinoma_skin, carcinoma_squamous, carcinoma_thyroid, carcinoma_urinary, chondrosarcoma, cns_tumor, eye_tumor, eyelid_tumor, gist, hair_follicle_tumor, hemangiosarcoma, histiocytic_sarcoma, leiomyosarcoma, leukemia, liposarcoma, liver_tumor, lymphoma, mammary_tumor, mast_cell_tumor, melanoma, nasal_tumor, nephroblastoma, osteosarcoma, perianal_adenoma, plasma_cell_neoplasm, rhabdomyosarcoma, skin_appendage_tumor, soft_tissue_sarcoma, spleen_tumor, testicular_tumor, thymoma, unknown_neoplasia."

## S2. Feature Engineering and Data Preprocessing

### S2.1. Complete List of Features.

The initial feature set comprised all available routine laboratory parameters from the CBC and serum biochemistry panels, along with basic demographic information. Two composite ratios, the Neutrophil-to-Lymphocyte Ratio (NLR) and Platelet-to-Lymphocyte Ratio (PLR), were engineered based on their established role as markers of systemic inflammation. The complete list of all extracted and engineered features is provided in Table S2.1.

### Table S2.1. Complete List of Initial, Engineered Features and the Manually Selected Feature Set.

This table includes *CBC and biochemistry features* extracted from the GRLS dataset, as well as the *engineered features* and *manually selected feature set* which contains 15 biomarkers selected for their established clinical relevance to common paraneoplastic syndromes, such as anemia, thrombocytopenia, hypercalcemia, hypoglycemia, elevated liver enzymes, and changes in white blood cell count.

| Category | Feature Name |
| --- | --- |

| Complete Blood Count (CBC) | Hematocrit (HCT) |
|---|---|
| | Hemoglobin (HGB) |
| | MCH |
| | MCHC |
| | MCV |
| | RBC |
| | WBC |
| | Band neutrophils |
| | Basophils |
| | Eosinophils |
| | Lymphocytes |
| | Monocytes |
| | Neutrophils |
| | Platelets |
| Biochemistry (Chem) | ALP |
| | ALT |
| | AST |
| | Albumin |
| | Globulin |
| | Albumin:Globulin Ratio |
| | Amylase |
| | BUN |
| | Creatinine |
| | BUN:Creatinine Ratio |
| | Calcium * |
| | Chloride |
| | Cholesterol |
| | Creatine Kinase (CK) |
| | GGT |
| | Glucose |
| | Lipase |
| | Magnesium |
| | Phosphorus |
| | Potassium |
| | Sodium |
| | Sodium:Potassium Ratio |
| | Total Bilirubin |
| | Total Protein |
| | Triglycerides |
| **Demographics** | age_at_visit |
| | sex_status |
| **Engineered Features** | Neutrophil-to-Lymphocyte Ratio (NLR) |
| | Platelet-to-Lymphocyte Ratio (PLR) |
| **Manually Selected Feature Set** | age_at_visit |
| | Hemoglobin (HGB) |

|   | Platelets |
|---|---|
|   | MCHC |
|   | WBC |
|   | Band neutrophils |
|   | Lymphocytes |
|   | Albumin |
|   | Globulin |
|   | Albumin:Globulin Ratio |
|   | Magnesium |
|   | Calcium |
|   | Sodium |
|   | GGT |
|   | Glucose |

*Total Calcium was measured according to the original GRLS study (Labadie et al., 2022).

**S2.2. Missing Data Analysis.**

Prior to imputation, the dataset contained missing values for several laboratory parameters. Features with more than 70% missing values were excluded from the analysis. The ten features with the highest percentage of missing values are summarized in Table S2.2. A visual overview of the missing data pattern across all features is provided in Figure S2.3.

**Table S2.2: Summary of Missing Data for Top 10 Features.**

| Features | Missing Count | Missing Percentage (%) |
|---|---|---|
| **Band neutrophils** | 13262 | 59.05 |
| **Lipase** | 13010 | 57.93 |
| **Amylase** | 3934 | 17.52 |
| **Triglycerides** | 3830 | 17.05 |
| **Basophils** | 3663 | 16.31 |
| **MCH** | 3635 | 16.18 |
| **Sodium:Potassium Ratio** | 326 | 1.45 |
| **Creatine Kinase (CK)** | 326 | 1.45 |
| **Albumin:Globulin Ratio** | 325 | 1.45 |
| **Globulin** | 325 | 1.45 |

**Figure S2.3: Missing Data Heatmap.**

The heatmap visualizes the occurrence of missing values (yellow) across all features and a random subset of visits.

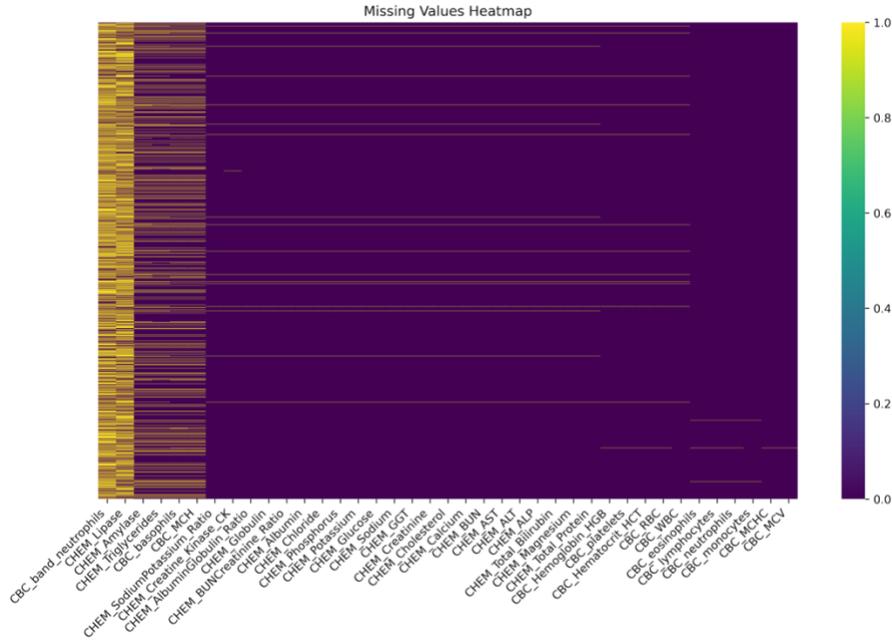

**Figure S2.4: Feature Correlation Analysis.**

The correlation between all features in the final analytical dataset was assessed. The heatmap shows the correlation matrix, which helps identify highly correlated variables (multicollinearity). Red indicates positive correlation, blue indicates negative correlation, and white indicates no correlation. The strength of the color corresponds to the magnitude of the correlation coefficient.

As expected, strong positive correlations are observed between related hematological parameters, such as RBC&HGB, HGB&HCT, RBC&HCT, WBC&Neutrophils, Creatinine&BUN, ect.

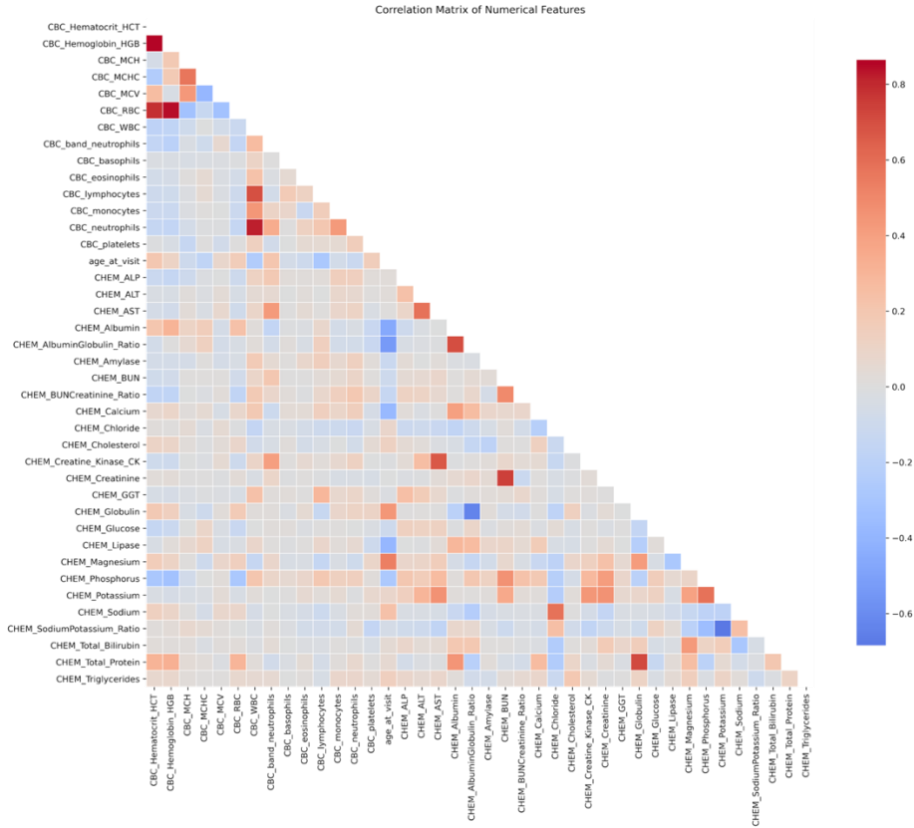

## S3. Model Development and Hyperparameter Tuning

### S3.1: Detailed Hyperparameter Search Space for All Models

Hyperparameter tuning was performed using 5-fold cross-validated GridSearchCV on the training set, with the Matthews Correlation Coefficient (MCC) as the optimization metric. The complete search space for each machine learning algorithm is detailed below. The specific optimized parameters for the top-performing models are reported in Table 3.2 of the main manuscript.

### Table S3.1: Comprehensive Hyperparameter Search Space for All Machine Learning Models

All combinations of the listed parameter values were evaluated during the grid search process.

| Base Model | Hyperparameter | Values Tested | Description |
|---|---|---|---|
| **Logistic Regression** | C | [0.001, 0.01, 0.1, 1, 10, 100] | Inverse of regularization strength; smaller values specify stronger regularization. |

| Model | Hyperparameter | Values | Description |
|---|---|---|---|
| **Random Forest** | n_estimators | [50, 100, 200] | Number of trees in the forest. |
| | max_depth | [3, 5, 10, 15, None] | Maximum depth of the tree. None indicates nodes are expanded until all leaves are pure. |
| | min_samples_split | [2, 5, 10] | Minimum number of samples required to split an internal node. |
| | min_samples_leaf | [1, 2, 4] | Minimum number of samples required to be at a leaf node. |
| **XGBoost** | n_estimators | [100, 200, 300] | Number of gradient boosted trees. |
| | max_depth | [3, 6, 9] | Maximum tree depth for base learners. |
| | learning_rate | [0.01, 0.1, 0.2] | Boosting learning rate (eta). |
| | subsample | [0.8, 0.9, 1.0] | Subsample ratio of the training instances. |
| **LightGBM** | n_estimators | [100, 200, 300] | Number of boosted trees to fit. |
| | max_depth | [3, 6, 9] | Maximum tree depth for base learners, -1 indicates no limit. |
| | learning_rate | [0.01, 0.1, 0.2] | Boosting learning rate. |
| **Multi-Layer Perceptron (MLP)** | hidden_layer_sizes | [(50,), (100,), (50, 50), (100, 50), (100, 50, 25)] | The ith element represents the number of neurons in the ith hidden layer. |
| | alpha | [0.0001, 0.001, 0.01, 0.1] | L2 penalty (regularization term) parameter. |
| | learning_rate | ['constant', 'adaptive'] | Learning rate schedule for weight updates. |
| **Naïve Bayes** | *None* | {} | No hyperparameters were tuned for this model. |